\documentclass[letterpaper, 10 pt, conference]{class/ieeeconf}  

\IEEEoverridecommandlockouts                              
\usepackage{amsmath}
\usepackage{amssymb}
\usepackage{latexsym}
\usepackage{url}
\usepackage{cite}
\usepackage{relsize}
\usepackage{multirow}
\usepackage{afterpage}
\usepackage{ifthen}
\usepackage{graphicx}
\usepackage{algpseudocode,algorithm,algorithmicx}
\usepackage{subfigure}
\usepackage[keeplastbox]{flushend}
\usepackage{epstopdf}
\usepackage{stackengine}
\usepackage{gensymb}
\usepackage{booktabs}
\usepackage{makecell}
\usepackage{hhline}
\usepackage{courier}
\usepackage{lipsum}
\usepackage{xcolor}
\usepackage{siunitx}

\title{\LARGE \bf Autonomous Navigation in Complex Environments with \\Deep Multimodal Fusion Network}
\author{Anh Nguyen$^1$, Ngoc Nguyen$^2$, Kim Tran$^2$, Erman Tjiputra$^2$, Quang D. Tran$^2$
\thanks{$^1$Department of Computing, Imperial College London, UK {\tt a.nguyen@imperial.ac.uk}}
\thanks{$^2$AIOZ Pte Ltd, Singapore {\tt quang.tran@aioz.io}}
}
\begin{document}

\newtheorem{problem}{Problem}
\newtheorem{lemma}{Lemma}
\newtheorem{theorem}[lemma]{Theorem}
\newtheorem{claim}{Claim}
\newtheorem{corollary}[lemma]{Corollary}
\newtheorem{definition}[lemma]{Definition}
\newtheorem{proposition}[lemma]{Proposition}
\newtheorem{remark}[lemma]{Remark}
\newenvironment{LabeledProof}[1]{\noindent{\it Proof of #1: }}{\qed}

\def\beq#1\eeq{\begin{equation}#1\end{equation}}
\def\bea#1\eea{\begin{align}#1\end{align}}
\def\beg#1\eeg{\begin{gather}#1\end{gather}}
\def\beqs#1\eeqs{\begin{equation*}#1\end{equation*}}
\def\beas#1\eeas{\begin{align*}#1\end{align*}}
\def\begs#1\eegs{\begin{gather*}#1\end{gather*}}

\newcommand{\poly}{\mathrm{poly}}
\newcommand{\eps}{\epsilon}
\newcommand{\e}{\epsilon}
\newcommand{\polylog}{\mathrm{polylog}}
\newcommand{\rob}[1]{\left( #1 \right)} 
\newcommand{\sqb}[1]{\left[ #1 \right]} 
\newcommand{\cub}[1]{\left\{ #1 \right\} } 
\newcommand{\rb}[1]{\left( #1 \right)} 
\newcommand{\abs}[1]{\left| #1 \right|} 
\newcommand{\zo}{\{0, 1\}}
\newcommand{\zonzo}{\zo^n \to \zo}
\newcommand{\zokzo}{\zo^k \to \zo}
\newcommand{\zot}{\{0,1,2\}}
\newcommand{\en}[1]{\marginpar{\textbf{#1}}}
\newcommand{\efn}[1]{\footnote{\textbf{#1}}}
\newcommand{\vecbm}[1]{\boldmath{#1}} 
\newcommand{\uvec}[1]{\hat{\vec{#1}}}
\newcommand{\thv}{\vecbm{\theta}}
\newcommand{\junk}[1]{}
\newcommand{\var}{\mathop{\mathrm{var}}}
\newcommand{\rank}{\mathop{\mathrm{rank}}}
\newcommand{\diag}{\mathop{\mathrm{diag}}}
\newcommand{\tr}{\mathop{\mathrm{tr}}}
\newcommand{\acos}{\mathop{\mathrm{acos}}}
\newcommand{\atantwo}{\mathop{\mathrm{atan2}}}
\newcommand{\SVD}{\mathop{\mathrm{SVD}}}
\newcommand{\quadf}{\mathop{\mathrm{q}}}
\newcommand{\linterp}{\mathop{\mathrm{l}}}
\newcommand{\sgn}{\mathop{\mathrm{sign}}}
\newcommand{\sym}{\mathop{\mathrm{sym}}}
\newcommand{\avg}{\mathop{\mathrm{avg}}}
\newcommand{\mean}{\mathop{\mathrm{mean}}}
\newcommand{\erf}{\mathop{\mathrm{erf}}}
\newcommand{\grad}{\nabla}
\newcommand{\R}{\mathbb{R}}
\newcommand{\defeq}{\triangleq}
\newcommand{\dims}[2]{[#1\!\times\!#2]}
\newcommand{\sdims}[2]{\mathsmaller{#1\!\times\!#2}}
\newcommand{\udims}[3]{#1}
\newcommand{\udimst}[4]{#1}
\newcommand{\com}[1]{\rhd\text{\emph{#1}}}
\newcommand{\ind}{\hspace{1em}}
\newcommand{\argmin}[1]{\underset{#1}{\operatorname{argmin}}}
\newcommand{\floor}[1]{\left\lfloor{#1}\right\rfloor}
\newcommand{\step}[1]{\vspace{0.5em}\noindent{#1}}
\newcommand{\quat}[1]{\ensuremath{\mathring{\mathbf{#1}}}}
\newcommand{\norm}[1]{\left\lVert#1\right\rVert}
\newcommand{\ignore}[1]{}
\newcommand{\specialcell}[2][c]{\begin{tabular}[#1]{@{}c@{}}#2\end{tabular}}
\newcommand*\Let[2]{\State #1 $\gets$ #2}
\newcommand{\algorithmicbreak}{\textbf{break}}
\newcommand{\Break}{\State \algorithmicbreak}
\newcommand{\ra}[1]{\renewcommand{\arraystretch}{#1}}

\renewcommand{\vec}[1]{\mathbf{#1}} 

\algdef{S}[FOR]{ForEach}[1]{\algorithmicforeach\ #1\ \algorithmicdo}
\algnewcommand\algorithmicforeach{\textbf{for each}}
\algrenewcommand\algorithmicrequire{\textbf{Require:}}
\algrenewcommand\algorithmicensure{\textbf{Ensure:}}
\algnewcommand\algorithmicinput{\textbf{Input:}}
\algnewcommand\INPUT{\item[\algorithmicinput]}
\algnewcommand\algorithmicoutput{\textbf{Output:}}
\algnewcommand\OUTPUT{\item[\algorithmicoutput]}

\maketitle
\thispagestyle{empty}
\pagestyle{empty}

\begin{abstract}
Autonomous navigation in complex environments is a crucial task in time-sensitive scenarios such as disaster response or search and rescue. However, complex environments pose significant challenges for autonomous platforms to navigate due to their challenging properties: constrained narrow passages, unstable pathway with debris and obstacles, or irregular geological structures and poor lighting conditions. In this work, we propose a multimodal fusion approach to address the problem of autonomous navigation in complex environments such as collapsed cites, or natural caves. We first simulate the complex environments in a physics-based simulation engine and collect a large-scale dataset for training. We then propose a Navigation Multimodal Fusion Network (NMFNet) which has three branches to effectively handle three visual modalities: laser, RGB images, and point cloud data. The extensively experimental results show that our NMFNet outperforms recent state of the art by a fair margin while achieving real-time performance. We further show that the use of multiple modalities is essential for autonomous navigation in complex environments. Finally, we successfully deploy our network to both simulated and real mobile robots.

\end{abstract}

\section{INTRODUCTION} \label{Sec:Intro}
Autonomous navigation is a long-standing field of robotics research, which provides an essential capability for mobile robot to execute a series of tasks on the same environments performed by human everyday. In general, the task of autonomous navigation is to control a robot navigate around the environment without colliding with obstacles. It can be seen that navigation is an elementary skill for intelligent agents, which requires decision-making across a diverse range of scales in time and space. In practice, autonomous navigation is not a trivial task since the robot needs to close the perception-control loop under the uncertainty in order to obtain the autonomy.

Recently, the learning-based approaches (e.g., deep learning models, etc.) have demonstrated the ability to directly derive end-to-end policies which map raw sensor data to control commands~\cite{pfeiffer2017perception, nguyen2019v2cnet}. This end-to-end approach also reduces the complexity of the implementation and effectively utilizes input data from different sensors (e.g., depth camera, laser) thereby reducing cost, power and computational time. One more advantage is that the end-to-end relationship between input data and control outputs can result in an arbitrarily non-linear complex model (i.e., sensor to actuation) which has yielded surprisingly encouraging results in different control problems such as lane following~\cite{Gurghian16}, autonomous driving~\cite{BojarskiTDFFGJM16}, and Unmanned Aerial Vehicles (UAV) control~\cite{Monajjemi16}. However, most of the previous works are only tested in man-made normal environments, while navigating in complex environments such as collapsed houses/cities suffered from a disaster (e.g. an earthquake) or a natural caves still remains as an open problem.


\begin{figure}[!t] 
    \centering
    \includegraphics[width=0.99\linewidth, height=0.82\linewidth]{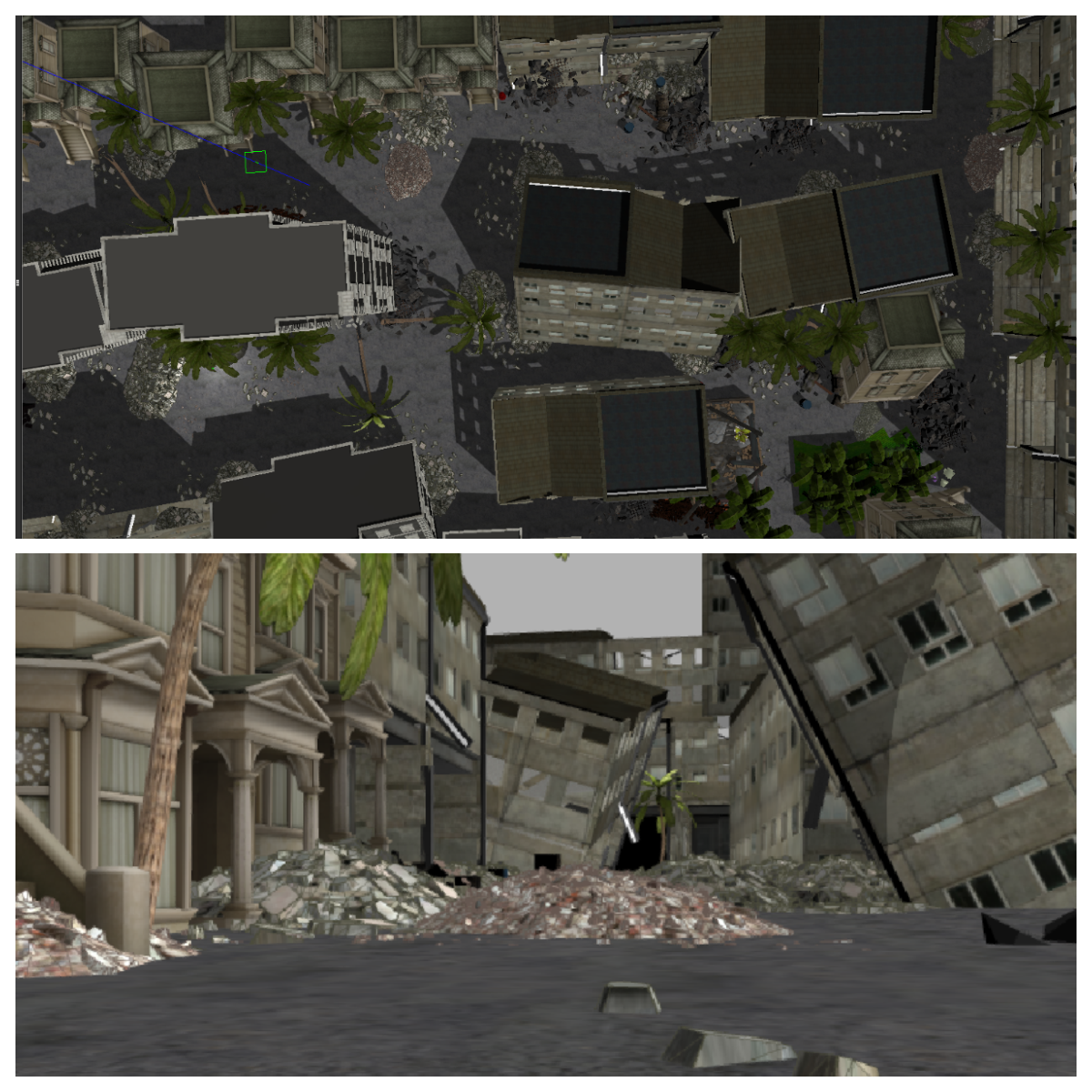} 
    \vspace{0ex}
    \caption{A visualization of a collapsed city, a complex environment from our simulation. \textbf{Top:} The collapsed city from a top view; \textbf{Bottom:} A mobile robot's view.}
    \label{Fig:intro} 
\end{figure}
 
Unlike normal environments (e.g., man-made road) which have clear visual clues in normal condition, complex environments such as collapsed cities or natural caves pose significant challenges for autonomous navigation~\cite{Howard2008}~\cite{piotr_2017_navigate_in_complex_environments}. The main reason is that complex environments usually have very challenging visual and physical properties. For example, the collapsed cities may have constrained narrow passages, vertical shafts, unstable pathways with debris and broken objects (Fig.~\ref{Fig:intro}); or the natural caves often have irregular geological structures, narrow passages, and poor lighting condition. Autonomous navigation with intelligent robots in complex environments, however, is a crucial task, especially in time-sensitive scenarios such as disaster response, search and rescue, etc.  Recently, the DARPA Subterranean Challenge~\cite{subt-challenge} was organized to explore novel methods for quickly mapping, navigating, and searching in complex underground environments such as human-made tunnel systems, urban underground, and natural cave networks. 

Inspired by the DARPA Subterranean Challenge, in this work, we propose a learning-based system for end-to-end mobile robot navigation in complex environments such as collapsed cities, collapsed houses, and natural caves. To overcome the difficulty in data collection, we build a large-scale simulation environment which allows us to collect the training data and deploy the learned control policy. We then propose a new Navigation Multimodal Fusion Network (NMFNet) that effectively learns the visual perception from sensor fusion and allows the robot to autonomously navigate in complex environments. To summarize, our main contributions are as follows:
\begin{itemize}
    \item We introduce new simulation models that can be used to record large-scale datasets for autonomous navigation in complex environments.
    \item We present a new deep learning method that fuses both laser data, 2D images, and 3D point cloud to improve the navigation ability of the robot in complex environments.
    \item We show that the use of multiple visual modalities is essential to learn a robust robot control policy for autonomous navigation in complex environments in order to deploy in real-world scenarios.    
\end{itemize}

The remainder of this paper is organized as follows: Section \ref{Sec:rw} discusses the related background. Section \ref{Sec_visual} presents the visual multimodal input used in our method. Section \ref{Sec_main_net} introduces our new multimodal fusion network and its architecture. In section \ref{Sec:exp}, we present our extensively experimental results. Section \ref{Sec:con} concludes the paper and discusses the future work.
\section{Related Work} \label{Sec:rw}
Multiple sensor fusion for autonomous robot navigation is a popular research topic in robotics~\cite{kam1997sensor}. Traditional methods  tackle this problem using algorithms based on Kalman Filter~\cite{dobrev2016multi}. The advantage of this approach is the ability to fuse data from different sensors and sensor types such as visual, inertial, GPS, or pressure sensors. Lynen et al.~\cite{lynen2013robust} proposed a method based on Extended Kalman Filter (EKF) for Micro Aerial Vehicle (MAV) navigation. In~\cite{du2020real}, the authors developed an algorithm based on EKF to estimated the state of an UAV in multi-environments in real-time. Mascaro et al.~\cite{mascaro2018gomsf} proposed a graph-optimization method to fuse data from multi sensors for UAV pose estimation. Apart from the traditional localization and navigation task, multimodal fusion is also used other applications such as object detection~\cite{mees16iros} or semantic segmentation~\cite{feng2020deep, valada2017adapnet} in changing environments. In both~\cite{mees16iros, valada2017adapnet} multimodal data from visual sensors are combined and learned in a deep learning framework to deal with challenging lighting conditions.

More recently, many methods have been proposed to directly learn control policies from raw sensory data. These methods can be divided into two main categories: reinforcement learning~\cite{duan2016benchmarking} and supervised learning~\cite{ross2013learning, XuGYD16,giusti2015machine}. With the rise of deep learning, Convolution Neural Networks (CNN) was widely used to train an end-to-end perception system~\cite{xu2017end, nguyen2020end, kim2017interpretable, nguyen2019object, richter2017safe, gao2017intention}. In~\cite{bojarski2016end2end_car}, Bojarski et al. proposed the first end-to-end navigation system for autonomous car using 2D images. Smolyanskiy et al.~\cite{smolyanskiy2017toward} extended this idea for flying robots using three cameras as the input. Similarly, the authors of DroNet~\cite{loquercio2018dronet} used CNN to learn the steering angle and predict the collision probability given the RGB image as the input. Gandhi et al.~\cite{gandhi2017learning} introduced a navigation method for UAV by learning from negative and positive crashing trials. In~\cite{amini2018variational}~\cite{amini2018learning}, CNN and Variational Autoencoder were combined to estimate the steering control signal. Monajjemi et al.~\cite{Monajjemi16} proposed a new method for agile UAV control. More recently, the authors in~\cite{alexander_2019_variational_end2end} proposed to combine the navigation map with visual input to learn a deterministic control signal.

Reinforcement learning algorithms have been widely used to learn general policies from robot experiences~\cite{duan2016benchmarking, tai2017virtual, schulman2015trust}. In~\cite{lillicrap2015continuous}, the authors introduced a continuous control framework using deep reinforcement learning. Zhu et al.~\cite{zhu2017target} addressed the target-driven navigation problem given an input picture of a target object. Wortsman et al.\cite{wortsman2018_RL} introduced a self-adaptive visual navigation system using meta-learning. The authors in~\cite{delbrouck2018_object_navigation} used semantic information and spatial relationships to let a robot navigate to target objects. In~\cite{muller2018teaching}, an end-to-end regression system was introduced for UAV racing in simulation. The authors in~\cite{sadeghi2016cad2rl}\cite{mancini2017toward} proposed to train the reinforcement policy in simulation environments, then transfer the learned policy to the real-world. In~\cite{andersson2017deep}~\cite{dosovitskiy2016learning}, the authors combined deep reinforcement learning with CNN in an effort to leverage the advantages of both techniques. Piotr et al.~\cite{piotr_2017_navigate_in_complex_environments} proposed a method with augmented memory to train autonomous agents to navigate within large and visually rich environments (complicated 3D mazes).

While reinforcement learning methods learn the general control policies with nice mathematical formulation, they require many trial and error experiments which are dangerous and not realistic in real safety-critical robotic platforms~\cite{delbrouck2018_object_navigation}~\cite{muller2018teaching}. On the order hand, supervised learning methods use pre-collected data to learn the control policies. The supervision data can be obtained from the real human expert trajectories~\cite{loquercio2018dronet, gandhi2017learning} or traditional controllers\cite{kahn2017plato}. This is time-consuming and costly but doable with the real robots. Therefore, the supervised learning approach is usually more favorable over the reinforcement learning method when working with real robot platforms. However, it is not trivial to handle the domain-shift between expert guidance and the real robot trajectories in supervised learning methods.

\begin{figure}[!ht] 
    \centering
    \includegraphics[width=0.99\linewidth, height=0.85\linewidth]{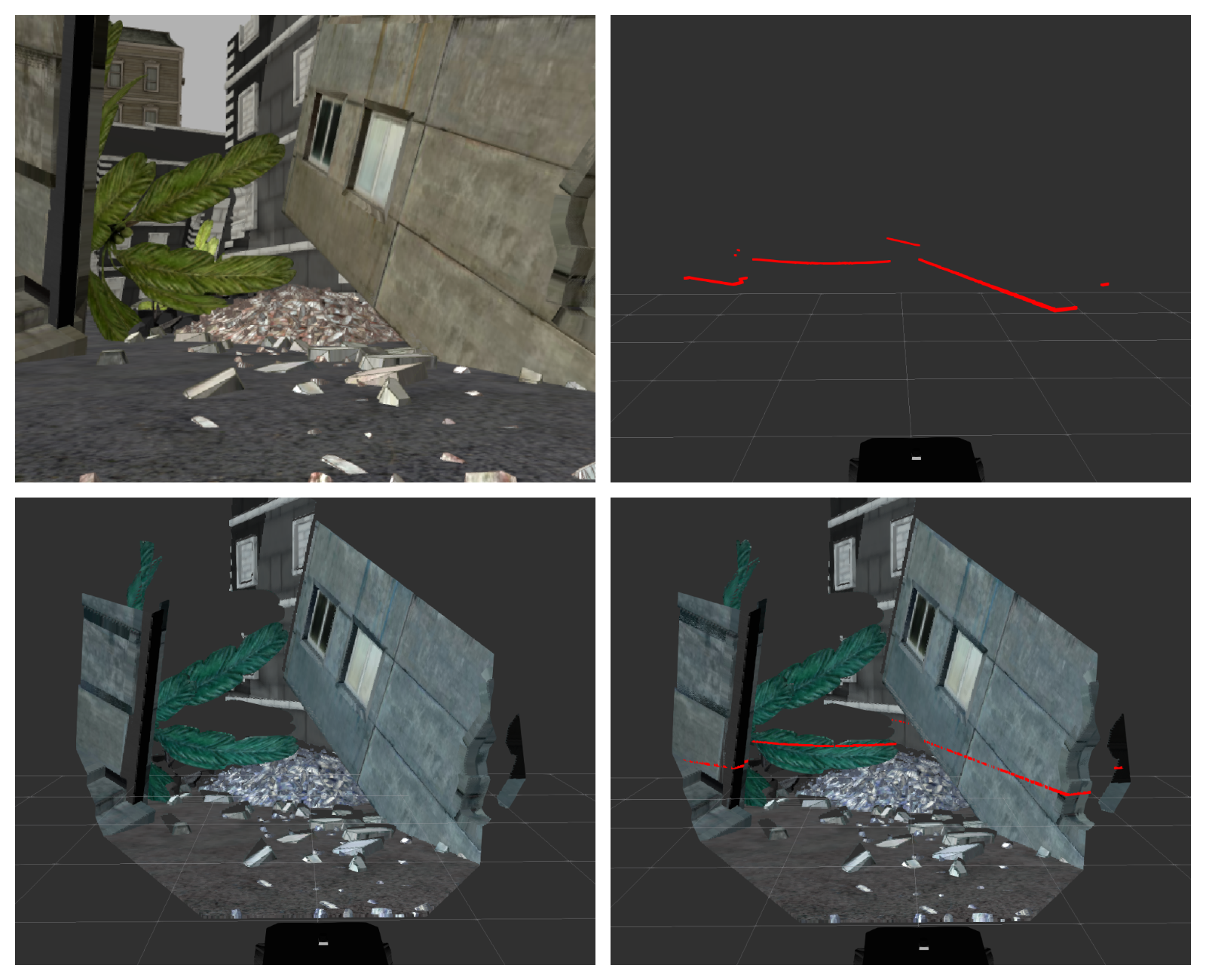} 
    \vspace{1ex}
    \caption{An illustration of three visual modalities used in our network. \textbf{Top row:} The RGB image (left) and the distance map from laser scanner (right). \textbf{Bottom row:} The 3D point cloud (left) and the overlay of the distance map in 3D point cloud (right).}
    \label{Fig_visual_input} 
\end{figure}

In this work, we choose the end-to-end supervised learning approach for the ease of deploying and testing in real robot systems. We first simulate the complex environments in physics-based simulation engine and collect a large-scale for supervised learning. We then proposed NMFNet, an effective deep learning framework to fuse visual input and allow the robots to navigate autonomously in complex environments.   
\section{Multimodal Input} \label{Sec_visual}
Complex environments such as natural cave networks or collapsed cities pose significant challenges for autonomous navigation due to their irregular structures, unexpected obstacles, and the poor lighting condition inside the environments. To overcome these natural difficulties, we use three visual input data in our method: RGB image $\mathcal{I}$, point cloud $\mathcal{P}$, and distance map $\mathcal{D}$ obtaining from the laser sensor. Intuitively, the use of all three visual modalities ensures that the robot's perception system has meaningful information from at least one modality during the navigation under challenging conditions such as lighting changes, sensor noise in depth channels due to reflective materials, or motion blur, etc. 

In practice, the RGB images and point clouds are captured using a depth camera mounted in front of the robot while the distance map is reconstructed from the laser data. In complex environments, while the RGB images and point clouds can provide the visual semantic information for the robot, the robot may need more useful information such as the distance map due to the presence of various obstacles. The distance map is reconstructed from the laser data as follows:


\begin{equation}
\begin{aligned}
x_i &= x_0 + d*cos(\pi - \phi * i) \\
y_i &= y_0 - d*sin(\phi * i)
\end{aligned}
\end{equation}
where $x_i,y_i$ is the coordinate of $i^{th}$ point on 2D distance map. $x_0,y_0$ is the coordinate of robot. $d$ is the distance from the laser sensor to the obstacle, and $\phi$ is the incremental angle of the laser sensor.

To keep the low latency between three visual modalities, we use only one laser scan to reconstruct the distance map. The scanning angle of the laser sensor is set to $180^\circ$ to cover the front view of the robot. This will help the robots aware of the obstacles from its far left/right hand side, since these obstacles may not be captured in the front camera which provides the RGB images and point cloud data. We notice that all three modalities are synchronized at each timestamp to ensure the robot is observing the same viewpoint at each control state. Fig.~\ref{Fig_visual_input} shows a visualization of three visual modalities used in our method.

\section{Methodology}\label{Sec_main_net}
\begin{figure*}[!ht] 
    \centering
    \includegraphics[width=0.99\linewidth, height=0.38428835489\linewidth]{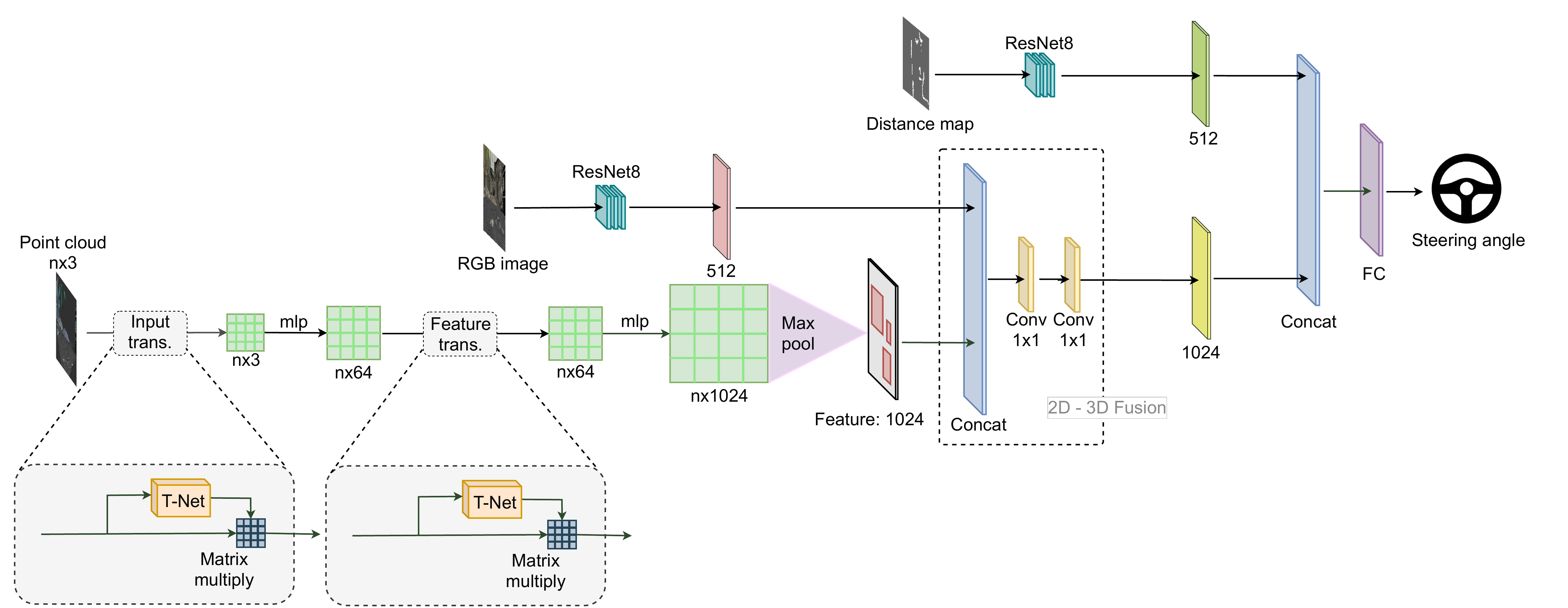} 
    \vspace{3ex}
    \caption{An overview of our NMFNet architecture. The network consists of three branches: The first branch learns the depth features from the distance map. The second branch extracts the features from RGB images, and the third branch handles the point cloud data. We then employ the 2D-3D fusion to predict the steering angle.}
    \label{Fig_method} 
\end{figure*}

As motivated by the recent trend in autonomous driving~\cite{smolyanskiy2017toward, loquercio2018dronet, alexander_2019_variational_end2end}, our goal is to build a framework that can directly map the input sensory data $\mathbf{X}=(\mathcal{D}, \mathcal{P}, \mathcal{I})$, to the output steering commands $\mathbf{Y}$. To this end, we design NMFNet with three branches to handle three visual modalities. The architecture of our network is illustrated in Fig.~\ref{Fig_method}.

\subsection{2D Features}
\begin{figure}[h] 
    \centering
    \includegraphics[width=0.99\linewidth, height=0.6\linewidth]{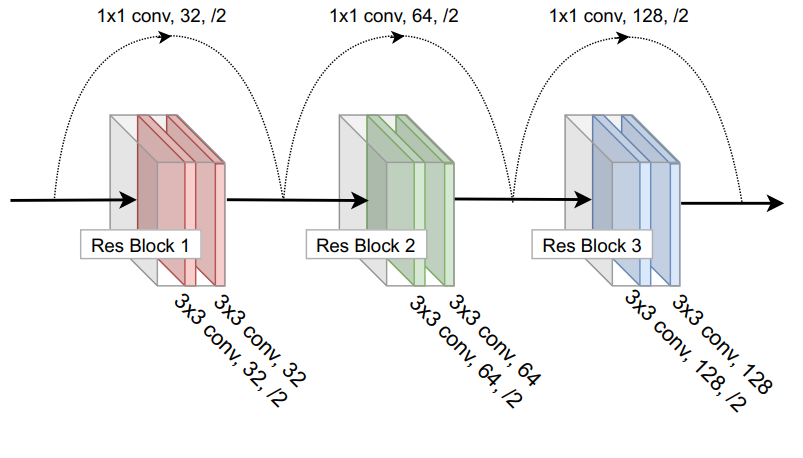} 
    \vspace{1ex}
    \caption{A detailed visualization of ResNet8.}
    \label{Fig_resnet8} 
\end{figure}

Learning meaningful features from 2D images is the key to success in many vision tasks. In this work, we use ResNet8 to extract deep features from the input RGB image and laser distance map. The ResNet8 has $3$ residual blocks, each block consists of a convolutional layer, ReLU, skip links and batch normalization operations. A detailed visualization of ResNet8 architecture can be found in Fig.~\ref{Fig_resnet8}. As in~\cite{loquercio2018dronet}, we choose ResNet8 to extract deep features from the 2D images since it is a light weight architecture and can achieve competitive performance while being robust again the vanishing/exploding gradient problems during training. 

\subsection{3D Point Cloud}  While the robot is navigating in complex environments, relying on 2D visual information maybe not enough. For example, the RGB images from the front camera of the robot are widely used in many end-to-end visual navigation systems~\cite{loquercio2018dronet, alexander_2019_variational_end2end}, however, in environments such as natural caves, the lighting condition can be a challenge to obtain clear visual images. Therefore, we propose to use point cloud as another visual input for autonomous navigation in complex environments. 

Specifically, we use the point cloud associated with the RGB images from the front camera of the robot. Although the point cloud from depth camera is ordered, it contains many points (e.g., in our camera setting, we have $640 \times 480 = 307, 200‬$ points in each cloud), including many missing points. In practice, due to the memory constraints, it is impractical to learn the geometric information from the huge point clouds~\cite{qi2017pointnet}. Therefore, we remove all the missing points and randomly select $20,480$ points to represent the cloud, hence the point cloud becomes unordered. The point cloud is expected to provide more geometry information of the environment for the network. However, extracting features from the unordered cloud is not a trivial task since the network needs to be invariant to all permutations of the input set.

To extract the point cloud feature vector, our network has to learn a model that is invariant to input permutation. As motivated by~\cite{qi2017pointnet}, we extract the features from the unordered point cloud by learning a symmetric function on transformed elements. Given an unordered point set $\{x_1, x_2, ..., x_n\}$ with $x_i \in \mathbb{R}^3$, we can define a symmetric function $f:X \to \mathbb{R}$ that maps a set of unordered points to a feature vector as follow:

\begin{equation}
f({x_1},{x_2},...,{x_n}) = \delta (\mathop {MAX}\limits_{i = 1,...,n} \{ \gamma ({x_i})\} )
\end{equation} 
where $MAX$ is a vector max operator that takes $n$ input vectors and returns a new vector of the element-wise maximum; $\delta$ and $\gamma$ are usually presented by neural networks.

In practice, $\delta$ and $\gamma$ function are approximated by an affine transformation matrix with a mini multi-layer preception network (i.e., T-net) and a matrix multiplication operation. Given the $n\times3$ unordered input points, we apply this transformation twice to learn the geometric feature from the pount cloud: input transformation and feature transformation. The input transformation uses raw point cloud as input and regresses to a $3 \times 3$ matrix. It consists of a three multi-layer perceptron network with layer output sizes are 64, 128, 1024, respectively. The output matrix is first initialized as an identity matrix and all layers have ReLU and batch normalization (except the last layer). We then feed the output of the first transformation to the second transformation network which has the same architecture and generates a $64 \times 64$ matrix as output. This matrix is also initialized as an identity and presents the learned features from the point cloud.

\subsection{Multimodal Fusion}
Given the features from the point cloud branch and the RGB image branch, we first do an early fusion by concatenating the features extracted from the input cloud with the deep features extracted from the RGB image. The intuition is that since both the RGB image and the point cloud are captured using a camera with the same viewpoint, fusing their features will let the robot aware of both visual information from RGB image and geometry clue from point cloud data. This concatenated feature is then fed through two $1\times1$ convolutional layers. Finally, we combine the features from 2D-3D fusion with the extracted features from the distance map branch. The steering angle is predicted from a final fully connected layer keeping all the features from the multimodal fusion network. 

\subsection{Training}
To train an end-to-end navigation network, two popular approaches are used: classification loss~\cite{amini2018learning} and regression loss\cite{loquercio2018dronet}. Methods use classification loss first bin the ground-truth steering angles into small and discrete groups, then learn possible controls as a classification problem using a softmax loss function. In practice, we have observed the instability during training due to the highly imbalanced statistic in the dataset. Therefore, we employ the regression loss to train the network end-to-end using the mean squared error (MSE) $L_2$ loss function between the ground-truth human actuated control, $y_i$, and the predicted control from the network $\hat{y}$:

\begin{equation}
L(y,\hat{y}) = \frac{1}{m}\sum\limits_{i = 1}^m {({y_i} - {{\hat y}_i})^2}
\end{equation}

Apart from training with normal data, we also employ the training method using domain randomisation~\cite{james2017transferring}. As shown in~\cite{james2017transferring}, this simple technique can effectively improve the generalization of the network when only simulation data are available for training.
\section{Experiments} \label{Sec:exp}
\subsection{Dataset}
\textbf{Data Collection} Unlike the traditional autonomous navigation problem for autonomous car or UAVs that can collect data in real-world setting~\cite{smolyanskiy2017toward, loquercio2018dronet, gandhi2017learning}, it is not a trivial task to build complex environments such as collapsed cities or a collapsed houses in real life. Therefore, we create the simulation models of these environments in Gazebo and collect the visual data from simulation. In particular, we collect the data from three types of complex environment: 
\begin{itemize}
\item Collapsed house: The house suffered from an accident or a disaster (e.g. an earthquake) with random objects on the ground.
\item Collapsed city: Similar to the collapsed house, but for the outdoor environment. In this scenario, the road has many debris from the collapsed house/wall. 
\item Natural cave: A long natural tunnel in a poor brightness condition with irregular geological structures.
\end{itemize}

To build the simulation environments, we first create the 3D model of normal daily objects in indoor and outdoor environments (e.g. beds, tables, lamps, computers, tools, trees, cars, rocks, etc.), including broken objects (e.g. broken vases, broken dishes, and debris). These objects are then manually chosen and placed in each environment to create the entire simulated environment. 

For each environment, we use a mobile robot model equipped with a laser sensor and a depth camera mounted on top of the robot to collect the visual data. The robot is controlled manually to navigate around each environment. We collect the visual data when the robot is moving. All the visual data $(\mathcal{D}, \mathcal{P}, \mathcal{I})$ are synchronized with a current steering signal of the robot at each timestamp. 

\textbf{Data Statistic}
In particular, we create $539$ 3D object models to build the complex environments. These objects are used to build $30$ environments in total (i.e., $10$ instances for each environment). In average, the collapsed house environments are built with approximately $130$ objects in an area of $400m^2$. The collapsed city has $275$ objects and spread in $3,000m^2$ while the natural cave environments are built with $60$ objects in approximately $4,000m^2$ area. We manually control the robot in $40$ hours to collect the data. 

\begin{figure}[h] 
    \centering
    \includegraphics[width=0.99\linewidth, height=0.7\linewidth]{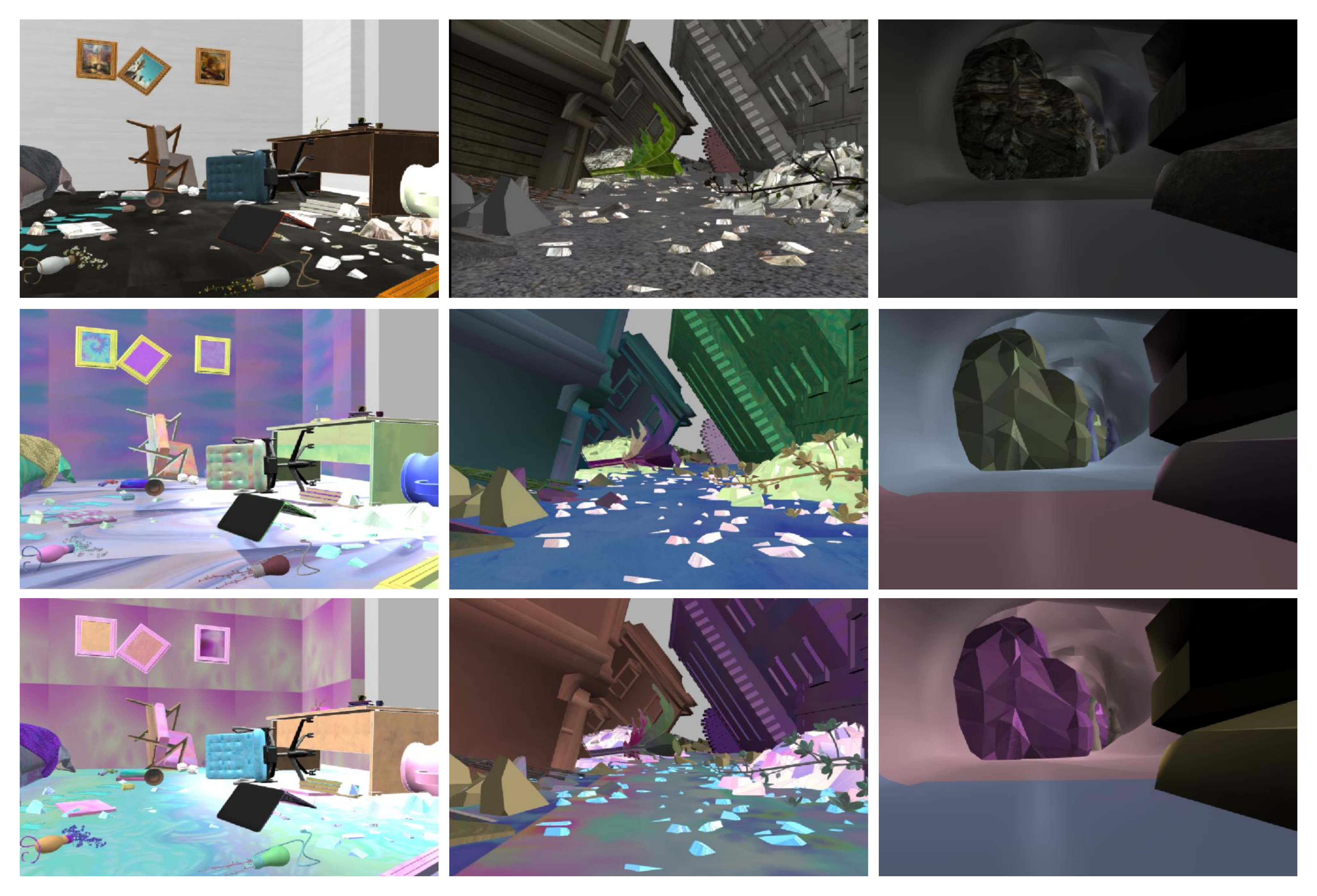} 
    \vspace{1ex}
    \caption{Different robot's views in our simulation of complex environments: collapsed city, natural cave, and collapsed house. \textbf{Top row:} RGB images from robot viewpoint in normal setting. \textbf{Other rows:} The same viewpoint when applying domain randomisation to the simulated environments.}
    \label{Fig_dataset} 
\end{figure}

In total, we collect around $40,000$ visual data triples ($\mathcal{D}, \mathcal{P}, \mathcal{I})$ for each environment type, resulting a large-scale dataset with $120,000$ records of synchronized RGB image, point cloud, laser distance map, and ground-truth steering angle. Around $45\%$ of the dataset are collected when we use domain randomisation by apply random texture to the environments (Fig.~\ref{Fig_dataset}). For each environment, we use $70\%$ data for training and $30\%$ data for testing.  All the 3D environments and our dataset will be made publicly available to encourage further research. 

\subsection{Implementation} We implement our network using Tensorflow framework~\cite{abadi2016tensorflow}. The network is optimized using stochastic gradient descent with the fix $0.01$ learning rate and $0.9$ momentum. The input RGB image and distance map size are ($480\times640$) and ($320\times640$), respectively, while the point cloud data are sampled to $20,480$ points. We train the network with the batch size of $8$ and the training time is approximately $30$ hours on an NVIDIA 2080 GPU.

\subsection{Baseline} We compare our method with the following recent state-of-the-art methods in autonomous navigation: DroNet~\cite{loquercio2018dronet}, VariationNet~\cite{amini2018variational}. We also present the result for Inception-V3 to serve as a baseline of deep architecture. We note that DroNet uses ResNet8 as the backbone to predict the steering angle and collision probability from RGB input. Since our dataset does not have collision ground-truth, we disable the collision probability branch of DroNet, and only use the regression branch to predict the result. Intuitively, DroNet architecture is similar to our RGB branch. All the baselines are trained with the data from domain randomisation. We show the results of our NMFNet under two settings: with domain randomisation (NMFNet with DR), and without using training data from domain randomisation (NMFNet without DR).

\subsection{Results}\label{sub_sec_results}
\begin{figure*}[!t]
  \centering
  
    \subfigure[Input Image]{\includegraphics[width=0.24\linewidth, height=0.19\linewidth]{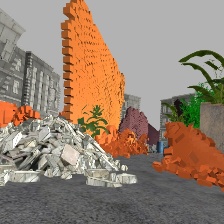}}
    \subfigure[RGB]{\includegraphics[width=0.24\linewidth, height=0.19\linewidth]{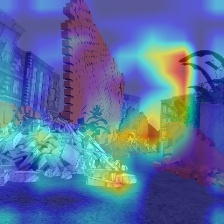}}
\subfigure[RGB + Point Cloud]{\includegraphics[width=0.24\linewidth, height=0.19\linewidth]{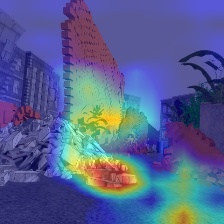}}
\subfigure[Fusion]{\includegraphics[width=0.24\linewidth, height=0.19\linewidth]{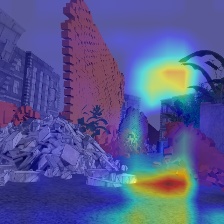}}

    \vspace{3ex}
    
 \caption{The activation map when different modalities are used to train the network. From left to right: \textbf{(a)} The input RGB image. \textbf{(b)} The activation map of the network uses only the RGB image as input. \textbf{(c)} The activation map of the network uses both RGB image and point cloud as input. \textbf{(d)} The activation map of the network uses fusion input (both RGB, point cloud and distance map). Overall, the network uses fusion input with all three modalities produces the most reliable heat map for navigation.}
 \label{Fig_activation}
\end{figure*}

Table~\ref{tb_result} summarizes the regression results using Root Mean Square Error (RMSE) of our NMFNet and other state-of-the-art methods. From the table, we can see that our NMFNet outperforms other methods by a significant margin. In particular, our NMFNet trained with domain randomisation data achieves $0.389$ RMSE which is a clear improvement over other methods using only RGB images such as DroNet~\cite{loquercio2018dronet}. This also confirms that using multi visual modalities input as in our fusion network is the key to successfully navigate in complex environments.

\begin{table}
\centering\ra{1.4}
\caption{RMSE Scores on the Test Set}
\label{tb_result}
\renewcommand\tabcolsep{4.5pt}

\hspace{1ex}

\begin{tabular}{@{}rccccc@{}}
\toprule 			&		
Input           &
House             & 
City             & 
Cave             & 
Average          \\     
\midrule
DroNet~\cite{loquercio2018dronet}				        & RGB         & 0.938 & 0.664   & 0.666   & 0.756   \\
Inception-V3~\cite{szegedy2016rethinking}			    & RGB         & 1.245 & 1.115   &1.121   & 1.16 \\
VariationNet~\cite{amini2018variational}                & RGB         & 1.510 & 1.290   & 1.507   & 1.436\\
\cline{1-6}
NMFNet without DR 	        & Fusion           & 0.482 & \textbf{0.365}   & 0.367   & 0.405 \\
NMFNet with DR 	        & Fusion           & \textbf{0.480} & \textbf{0.365}   & \textbf{0.321}   & \textbf{0.389}\\

\bottomrule
\end{tabular}
\end{table}


Within three complex environment types, we observe that the RMSE of the collapsed house results are higher than the collapsed city and the natural cave. A possible reason is that the collapsed house environment is much smaller than others while having more objects. Therefore, it would be more challenging for the robot to navigate in the collapsed house without colliding with the objects. From Table~\ref{tb_result}, we also notice that by employing domain randomisation, our NMFNet with DR shows a good improvement in comparison with the setting without domain randomisation (NMFNet without DR). On the other hand, the setup with VariationNet approach~\cite{amini2018variational} has the highest error in all three complex environments while the Inception-V3 shows reasonable results. 

\subsection{Contribution of Visual Modalities}
To understand the contribution of each modality to the results, we perform the following experiment: We first train a network that uses only a single modality (either RGB, distance map, or point cloud) as the input. Technically, each network in this experiment is a branch of our NMFNet. We then perform similar experiments using networks with two branches (i.e., RGB + distance map, RGB + point cloud, and distance map + point cloud) as the input. All the networks use the training set with extra data from domain randomisation.

Table~\ref{tb_modality} shows the RMSE scores when different modalities are used to train the system. We first notice that the network that uses only point cloud data as the input does not converge. This confirms that learning meaningful features from point cloud data is challenging, especially in complex environments. On the other hand, we achieve a surprisingly good result when the distance map modality is used as the input. The other combinations between two modalities show reasonable accuracy, however, we achieve the best result when the network is trained end-to-end using the fusion from all three modalities: rgb, distance map from laser camera, and point cloud from the depth camera. To further verify the contribution of each modality, we employ Grad-CAM~\cite{selvaraju2017grad} to visualize the activation map of the network when different modality is used. Fig.~\ref{Fig_activation} shows the qualitative visualization under three input settings: RGB, RGB + point cloud, and fusion. From Fig.~\ref{Fig_activation}, we can see that from a same viewpoint, the network that uses fusion data makes the most logical decision since its attention lays on feasible regions for navigation, while other networks trained with only RGB image or RGB + point cloud show more noisy attention.

We also note that the inference time of our NMFNet is approximately 100ms on an NVIDIA 2080 GPU. This allows our method to be used in a wide range of robotic applications. More qualitative results including the deployment of NMFNet on BeetleBot~\cite{nguyenbeetlebot} can be found at \url{https://sites.google.com/site/multimodalnavigation/}.

\begin{table}
\centering\ra{1.4}
\caption{RMSE Scores of Networks using Different Modality}
\label{tb_modality}
\renewcommand\tabcolsep{4.5pt}

\hspace{1ex}

\begin{tabular}{@{}rccccc@{}}
\toprule 			&	
                
Input           &
House             & 
City             & 
Cave             & 
Average          \\     
\midrule
& RGB                  			& 0.938 & 0.664   & 0.666   & 0.756   \\
& Distance Map         			& 0.730 & 0.579   & 0.602   & 0.637    \\
& Point Cloud               	&   - 	&   -  	  &   -    	&   -    \\
& RGB + Point Cloud         	& 0.718 & 0.499   & 0.783   & 0.667   \\
& RGB + Distance Map         	& 0.568 & 0.452   & 0.503   & 0.508   \\

& Distance Map + Point Cloud    & 0.631 & 0.474   & 0.592   & 0.566   \\

\cline{1-6}
& Fusion           				& 0.480 & 0.365   & 0.321   & 0.389\\

\bottomrule
\end{tabular}
\end{table}

\section{Conclusions and Future Work}\label{Sec:con}
We propose NMFNet, an end-to-end and real-time deep learning framework for autonomous navigation in complex environments. Our network has three branches and effectively learns the visual fusion input data. Furthermore, we show that the use of mutilmodal sensor data is essential for autonomous navigation in complex environments. The extensively experimental results show that our NMFNet outperforms recent state-of-the-art methods by a fair margin while achieving real-time performance and generalizing well on unseen environments. 

Currently, our NMFNet shows limitation on scenarios when the robot has to cross small debris or obstacles. In the future, we would like to quantitatively evaluate and address this problem. Another interesting direction is to combine our method with uncertainty estimation~\cite{richter2017safe} or a goal-driven navigation task~\cite{gao2017intention} for more wide-range of applications.


\section*{}
\addcontentsline{toc}{section}{}

\bibliographystyle{class/IEEEtran}
\bibliography{class/IEEEabrv,class/reference}

\begin{thebibliography}{10}
\providecommand{\url}[1]{#1}
\csname url@rmstyle\endcsname
\providecommand{\newblock}{\relax}
\providecommand{\bibinfo}[2]{#2}
\providecommand\BIBentrySTDinterwordspacing{\spaceskip=0pt\relax}
\providecommand\BIBentryALTinterwordstretchfactor{4}
\providecommand\BIBentryALTinterwordspacing{\spaceskip=\fontdimen2\font plus
\BIBentryALTinterwordstretchfactor\fontdimen3\font minus
  \fontdimen4\font\relax}
\providecommand\BIBforeignlanguage[2]{{%
\expandafter\ifx\csname l@#1\endcsname\relax
\typeout{** WARNING: IEEEtran.bst: No hyphenation pattern has been}%
\typeout{** loaded for the language `#1'. Using the pattern for}%
\typeout{** the default language instead.}%
\else
\language=\csname l@#1\endcsname
\fi
#2}}

\bibitem{pfeiffer2017perception}
M.~Pfeiffer, M.~Schaeuble, J.~Nieto, R.~Siegwart, and C.~Cadena, ``From
  perception to decision: A data-driven approach to end-to-end motion planning
  for autonomous ground robots,'' in \emph{ICRA}, 2017.

\bibitem{nguyen2019v2cnet}
A.~Nguyen, T.-T. Do, I.~Reid, D.~G. Caldwell, and N.~G. Tsagarakis, ``V2cnet: A
  deep learning framework to translate videos to commands for robotic
  manipulation,'' \emph{arXiv preprint arXiv:1903.10869}, 2019.

\bibitem{Gurghian16}
A.~{Gurghian}, T.~{Koduri}, S.~V. {Bailur}, K.~J. {Carey}, and V.~N. {Murali},
  ``Deeplanes: End-to-end lane position estimation using deep neural
  networks,'' in \emph{CVPR}, 2016.

\bibitem{BojarskiTDFFGJM16}
M.~Bojarski, D.~D. Testa, D.~Dworakowski, B.~Firner, B.~Flepp, P.~Goyal, L.~D.
  Jackel, M.~Monfort, U.~Muller, J.~Zhang, X.~Zhang, J.~Zhao, and K.~Zieba,
  ``End to end learning for self-driving cars.'' \emph{CoRR}, 2016.

\bibitem{Monajjemi16}
M.~{Monajjemi}, S.~{Mohaimenianpour}, and R.~{Vaughan}, ``Uav, come to me:
  End-to-end, multi-scale situated hri with an uninstrumented human and a
  distant uav,'' in \emph{IROS}, 2016.

\bibitem{Howard2008}
T.~Howard, C.~Green, D.~Ferguson, and A.~Kelly, ``State space sampling of
  feasible motions for high-performance mobile robot navigation in complex
  environments,'' \emph{Journal of Field Robotics}, 2008.

\bibitem{piotr_2017_navigate_in_complex_environments}
P.~Mirowski, R.~Pascanu, F.~Viola, H.~Soyer, A.~J. Ballard, A.~Banino,
  M.~Denil, R.~Goroshin, L.~Sifre, K.~Kavukcuoglu, \emph{et~al.}, ``Learning to
  navigate in complex environments,'' \emph{arXiv:1611.03673}, 2017.

\bibitem{subt-challenge}
``Darpa subterranean challenge,'' \url{https://www.subtchallenge.com/}.

\bibitem{kam1997sensor}
M.~Kam, X.~Zhu, and P.~Kalata, ``Sensor fusion for mobile robot navigation,''
  \emph{Proceedings of the IEEE}, 1997.

\bibitem{dobrev2016multi}
Y.~Dobrev, S.~Flores, and M.~Vossiek, ``Multi-modal sensor fusion for indoor
  mobile robot pose estimation,'' in \emph{2016 IEEE/ION Position, Location and
  Navigation Symposium (PLANS)}, 2016.

\bibitem{lynen2013robust}
S.~Lynen, M.~W. Achtelik, S.~Weiss, M.~Chli, and R.~Siegwart, ``A robust and
  modular multi-sensor fusion approach applied to mav navigation,'' in
  \emph{IROS}, 2013.

\bibitem{du2020real}
H.~Du, W.~Wang, C.~Xu, R.~Xiao, and C.~Sun, ``Real-time onboard 3d state
  estimation of an unmanned aerial vehicle in multi-environments using
  multi-sensor data fusion,'' \emph{Sensors}, 2020.

\bibitem{mascaro2018gomsf}
R.~Mascaro, L.~Teixeira, T.~Hinzmann, R.~Siegwart, and M.~Chli, ``Gomsf:
  Graph-optimization based multi-sensor fusion for robust uav pose
  estimation,'' in \emph{ICRA}, 2018.

\bibitem{mees16iros}
O.~Mees, A.~Eitel, and W.~Burgard, ``Choosing smartly: Adaptive multimodal
  fusion for object detection in changing environments,'' in \emph{IROS}, 2016.

\bibitem{feng2020deep}
D.~Feng, C.~Haase-Sch{\"u}tz, L.~Rosenbaum, H.~Hertlein, C.~Glaeser, F.~Timm,
  W.~Wiesbeck, and K.~Dietmayer, ``Deep multi-modal object detection and
  semantic segmentation for autonomous driving: Datasets, methods, and
  challenges,'' \emph{IEEE Transactions on Intelligent Transportation Systems},
  2020.

\bibitem{valada2017adapnet}
A.~Valada, J.~Vertens, A.~Dhall, and W.~Burgard, ``Adapnet: Adaptive semantic
  segmentation in adverse environmental conditions,'' in \emph{ICRA}, 2017.

\bibitem{duan2016benchmarking}
Y.~Duan, X.~Chen, R.~Houthooft, J.~Schulman, and P.~Abbeel, ``Benchmarking deep
  reinforcement learning for continuous control,'' in \emph{ICML}, 2016.

\bibitem{ross2013learning}
S.~Ross, N.~Melik-Barkhudarov, K.~S. Shankar, A.~Wendel, D.~Dey, J.~A. Bagnell,
  and M.~Hebert, ``Learning monocular reactive uav control in cluttered natural
  environments,'' in \emph{ICRA}, 2013.

\bibitem{XuGYD16}
H.~Xu, Y.~Gao, F.~Yu, and T.~Darrell, ``End-to-end learning of driving models
  from large-scale video datasets,'' \emph{CoRR}, 2016.

\bibitem{giusti2015machine}
A.~Giusti, J.~Guzzi, D.~C. Cire{\c{s}}an, F.-L. He, J.~P. Rodr{\'\i}guez,
  F.~Fontana, M.~Faessler, C.~Forster, J.~Schmidhuber, G.~Di~Caro,
  \emph{et~al.}, ``A machine learning approach to visual perception of forest
  trails for mobile robots,'' \emph{RA-L}, 2015.

\bibitem{xu2017end}
H.~Xu, Y.~Gao, F.~Yu, and T.~Darrell, ``End-to-end learning of driving models
  from large-scale video datasets,'' in \emph{CVPR}, 2017.

\bibitem{nguyen2020end}
A.~Nguyen, D.~Kundrat, G.~Dagnino, W.~Chi, M.~E. Abdelaziz, Y.~Guo, Y.~Ma,
  T.~M. Kwok, C.~Riga, and G.-Z. Yang, ``End-to-end real-time catheter
  segmentation with optical flow-guided warping during endovascular
  intervention,'' \emph{arXiv preprint arXiv:2006.09117}, 2020.

\bibitem{kim2017interpretable}
J.~Kim and J.~Canny, ``Interpretable learning for self-driving cars by
  visualizing causal attention,'' in \emph{ICCV}, 2017.

\bibitem{nguyen2019object}
A.~Nguyen, Q.~D. Tran, T.-T. Do, I.~Reid, D.~G. Caldwell, and N.~G. Tsagarakis,
  ``Object captioning and retrieval with natural language,'' in \emph{CVPRW},
  2019.

\bibitem{richter2017safe}
C.~Richter and N.~Roy, ``Safe visual navigation via deep learning and novelty
  detection,'' in \emph{RSS}, 2017.

\bibitem{gao2017intention}
W.~Gao, D.~Hsu, W.~S. Lee, S.~Shen, and K.~Subramanian, ``Intention-net:
  Integrating planning and deep learning for goal-directed autonomous
  navigation,'' \emph{arXiv:1710.05627}, 2017.

\bibitem{bojarski2016end2end_car}
M.~Bojarski, D.~Del~Testa, D.~Dworakowski, B.~Firner, B.~Flepp, P.~Goyal, L.~D.
  Jackel, M.~Monfort, U.~Muller, J.~Zhang, \emph{et~al.}, ``End to end learning
  for self-driving cars,'' \emph{arXiv:1604.07316}, 2016.

\bibitem{smolyanskiy2017toward}
N.~Smolyanskiy, A.~Kamenev, J.~Smith, and S.~Birchfield, ``Toward low-flying
  autonomous mav trail navigation using deep neural networks for environmental
  awareness,'' in \emph{IROS}, 2017.

\bibitem{loquercio2018dronet}
A.~Loquercio, A.~I. Maqueda, C.~R. del Blanco, and D.~Scaramuzza, ``Dronet:
  Learning to fly by driving,'' \emph{RA-L}, 2018.

\bibitem{gandhi2017learning}
D.~Gandhi, L.~Pinto, and A.~Gupta, ``Learning to fly by crashing,'' in
  \emph{IROS}, 2017.

\bibitem{amini2018variational}
A.~Amini, W.~Schwarting, G.~Rosman, B.~Araki, S.~Karaman, and D.~Rus,
  ``Variational autoencoder for end-to-end control of autonomous driving with
  novelty detection and training de-biasing,'' in \emph{IROS}, 2018.

\bibitem{amini2018learning}
A.~Amini, L.~Paull, T.~Balch, S.~Karaman, and D.~Rus, ``Learning steering
  bounds for parallel autonomous systems,'' in \emph{ICRA}, 2018.

\bibitem{alexander_2019_variational_end2end}
S.~K. Alexander~Amini, Guy~Rosman and D.~Rus, ``Variational end-to-end
  navigation and localization,'' \emph{arXiv:1811.10119v2}, 2019.

\bibitem{tai2017virtual}
L.~Tai, G.~Paolo, and M.~Liu, ``Virtual-to-real deep reinforcement learning:
  Continuous control of mobile robots for mapless navigation,'' in \emph{IROS},
  2017.

\bibitem{schulman2015trust}
J.~Schulman, S.~Levine, P.~Abbeel, M.~Jordan, and P.~Moritz, ``Trust region
  policy optimization,'' in \emph{ICML}, 2015.

\bibitem{lillicrap2015continuous}
T.~P. Lillicrap, J.~J. Hunt, A.~Pritzel, N.~Heess, T.~Erez, Y.~Tassa,
  D.~Silver, and D.~Wierstra, ``Continuous control with deep reinforcement
  learning,'' \emph{arXiv:1509.02971}, 2015.

\bibitem{zhu2017target}
Y.~Zhu, R.~Mottaghi, E.~Kolve, J.~J. Lim, A.~Gupta, L.~Fei-Fei, and A.~Farhadi,
  ``Target-driven visual navigation in indoor scenes using deep reinforcement
  learning,'' in \emph{ICRA}, 2017.

\bibitem{wortsman2018_RL}
M.~Wortsman, K.~Ehsani, M.~Rastegari, A.~Farhadi, and R.~Mottaghi, ``Learning
  to learn how to learn: Self-adaptive visual navigation using meta-learning,''
  \emph{arXiv:1812.00971}, 2018.

\bibitem{delbrouck2018_object_navigation}
J.-B. Delbrouck and S.~Dupont, ``Object-oriented targets for visual navigation
  using rich semantic representations,'' \emph{arXiv:1811.09178}, 2018.

\bibitem{muller2018teaching}
M.~Muller, V.~Casser, N.~Smith, D.~L. Michels, and B.~Ghanem, ``Teaching uavs
  to race: End-to-end regression of agile controls in simulation,'' in
  \emph{ECCV}, 2018.

\bibitem{sadeghi2016cad2rl}
F.~Sadeghi and S.~Levine, ``Cad2rl: Real single-image flight without a single
  real image,'' \emph{arXiv:1611.04201}, 2016.

\bibitem{mancini2017toward}
M.~Mancini, G.~Costante, P.~Valigi, T.~A. Ciarfuglia, J.~Delmerico, and
  D.~Scaramuzza, ``Toward domain independence for learning-based monocular
  depth estimation,'' \emph{RA-L}, 2017.

\bibitem{andersson2017deep}
O.~Andersson, M.~Wzorek, and P.~Doherty, ``Deep learning quadcopter control via
  risk-aware active learning,'' in \emph{AAAI}, 2017.

\bibitem{dosovitskiy2016learning}
A.~Dosovitskiy and V.~Koltun, ``Learning to act by predicting the future,''
  \emph{arXiv:1611.01779}, 2016.

\bibitem{kahn2017plato}
G.~Kahn, T.~Zhang, S.~Levine, and P.~Abbeel, ``Plato: Policy learning using
  adaptive trajectory optimization,'' in \emph{ICRA}, 2017.

\bibitem{qi2017pointnet}
C.~R. Qi, H.~Su, K.~Mo, and L.~J. Guibas, ``Pointnet: Deep learning on point
  sets for 3d classification and segmentation,'' in \emph{CVPR}, 2017.

\bibitem{james2017transferring}
S.~James, A.~J. Davison, and E.~Johns, ``Transferring end-to-end visuomotor
  control from simulation to real world for a multi-stage task,''
  \emph{arXiv:1707.02267}, 2017.

\bibitem{abadi2016tensorflow}
M.~Abadi, P.~Barham, J.~Chen, Z.~Chen, A.~Davis, J.~Dean, M.~Devin,
  S.~Ghemawat, G.~Irving, M.~Isard, \emph{et~al.}, ``Tensorflow: A system for
  large-scale machine learning,'' in \emph{Symposium on Operating Systems
  Design and Implementation}, 2016.

\bibitem{szegedy2016rethinking}
C.~Szegedy, V.~Vanhoucke, S.~Ioffe, J.~Shlens, and Z.~Wojna, ``Rethinking the
  inception architecture for computer vision,'' in \emph{CVPR}, 2015.

\bibitem{selvaraju2017grad}
R.~R. Selvaraju, M.~Cogswell, A.~Das, R.~Vedantam, D.~Parikh, and D.~Batra,
  ``Grad-cam: Visual explanations from deep networks via gradient-based
  localization,'' in \emph{ICCV}, 2017.

\bibitem{nguyenbeetlebot}
A.~Nguyen, E.~Tjiputra, and Q.~D. Tran, ``Beetlebot: A multi-purpose ai-driven
  mobile robot for realistic environments,'' in \emph{Proceedings of UKRAS20
  Conference: “Robots into the real world”}, 2020.

\end{thebibliography}
   
\end{document}